\documentclass{article}

\makeatletter
\def\input@path{{styles/}}
\makeatother

\usepackage{iclr2027_conference,times}

\usepackage{amsmath,amsfonts,bm}

\def\eqref#1{equation~\ref{#1}}

\def\1{\bm{1}}

\DeclareMathAlphabet{\mathsfit}{\encodingdefault}{\sfdefault}{m}{sl}
\SetMathAlphabet{\mathsfit}{bold}{\encodingdefault}{\sfdefault}{bx}{n}

\usepackage{booktabs}
\usepackage{multirow}
\usepackage{graphicx}
\usepackage{float}
\usepackage{wrapfig}
\usepackage{algorithm}
\usepackage{algpseudocode}
\usepackage{placeins}
\usepackage[table]{xcolor}
\usepackage{array}
\usepackage{hyperref}
\usepackage{url}
\usepackage{soul}

\definecolor{TablePrimary}{gray}{0.94}
\definecolor{TableGroup}{gray}{0.965}
\definecolor{TableMuted}{gray}{0.52}

\newcommand{\EvalTablePanelTitle}[1]{%
  \par\addvspace{5pt}%
  \textbf{#1}\par
  \vspace{2.5pt}%
}

\newcommand{\EvalTableMainStyle}{%
  \footnotesize
  \setlength{\tabcolsep}{4.0pt}%
  \renewcommand{\arraystretch}{1.08}%
}
\newcommand{\EvalTableStandardStyle}{%
  \scriptsize
  \setlength{\tabcolsep}{3.2pt}%
  \renewcommand{\arraystretch}{1.02}%
}
\newcommand{\EvalTableDenseStyle}{%
  \scriptsize
  \setlength{\tabcolsep}{3.0pt}%
  \renewcommand{\arraystretch}{0.82}%
}

\graphicspath{{assets/figures/}}

\title{EMAS: Stabilizing Multi-Agent System Evolution through Evidence-Guided Revision}

\author{%
  \makebox[\dimexpr\textwidth-2\tabcolsep\relax][c]{%
    \begin{tabular}{@{}c@{}}
      \normalfont
      \textbf{Chao Fei}\textsuperscript{1}\quad
      \textbf{Qingyi Si}\textsuperscript{2}\quad
      \textbf{Kaihua Liang}\textsuperscript{1}\quad
      \textbf{Yanghua Xiao}\textsuperscript{3}\quad
      \textbf{Panos Kalnis}\textsuperscript{1}\quad
      \textbf{Hongcheng Guo}\textsuperscript{3}
      \\[4pt]
      \normalfont\small
      \textsuperscript{1}King Abdullah University of Science and Technology (KAUST)\quad
      \textsuperscript{2}JD.com\quad
      \textsuperscript{3}Fudan University
    \end{tabular}%
  }%
}

\iclrfinalcopy

\begin{document}

\maketitle

\AddToShipoutPictureFG*{%
  \AtTextLowerLeft{%
    \put(0,-\LenToUnit{8pt}){%
      \begin{minipage}[t]{\textwidth}
        \rule{0.4\textwidth}{0.4pt}\\[-1pt]
        \footnotesize
        Code:
        \href{https://github.com/cf3i/Evolving-Multi-Agent-System}%
          {\nolinkurl{github.com/cf3i/Evolving-Multi-Agent-System}}.
      \end{minipage}%
    }%
  }%
}

\lhead{Preprint}

\begin{abstract}
Many methods for automated multi-agent system design optimize prompts and topologies during an initial design stage and then deploy the resulting system unchanged on subsequent samples.
Experience from these samples is rarely consolidated into reusable system updates, while accuracy-oriented designs may incur high token costs.
We introduce EMAS (Evolving Multi-Agent System), which uses this experience to revise MAS topology and prompts without updating LLM parameters, either to improve accuracy or to reduce cost.
EMAS converts traces into structured diagnoses that specify a revision operation and target.
It generates a candidate revision only when the same diagnosis recurs across samples and applies it only if paired validation against the current MAS meets the corresponding acceptance criterion.
Across four benchmarks and two LLMs, EMAS attains the highest task-weighted overall accuracy for both backbones and is best or tied in six of eight model--benchmark settings.
Within two evolution epochs, EMAS achieves relative gains of 6.30\% and 20.10\% in task-weighted accuracy on Kimi-K2-6 and Qwen3.6-27B, respectively.
On MBPP with Qwen3.6-27B, EMAS raises accuracy from 55.09\% to 89.12\% while reducing token use per task by 62.2\%.
These results show that EMAS can turn experience from new samples into reusable updates to MAS topology and prompts.
\end{abstract}

\section{Introduction}
\label{sec:introduction}

The goal of recursive self-improvement (RSI) is to build AI systems that improve the reusable mechanisms shaping their future behavior over time \citep{zelikman2024stop,zhang2026dgm}.
For an LLM-based multi-agent system (MAS), these mechanisms include more than model weights.
Prompts assign responsibilities, computation decomposes work, and topology routes intermediate artifacts.
Together, they form an executable system layer around the LLM.
An MAS can therefore evolve at the system level by revising its prompts and topology while keeping the underlying LLM fixed.
Experience collected as the MAS processes new tasks can provide evidence for these revisions.

Prior work has explored how to optimize this system layer.
GPTSwarm represents agentic systems as computational graphs and optimizes prompts and graph connectivity \citep{zhuge2024gptswarm}.
Automated design methods search over workflows and agent programs \citep{hu2025adas,zhang2025aflow}.
Symbolic learning methods treat prompts, tools, and their composition as learnable objects \citep{ou2025symboliclearning,wang2024semanticbackprop}.
Trace-conditioned methods show that execution evidence can guide revisions and that candidate updates can be screened empirically \citep{cheng2024trace,wang2024semanticbackprop}.
The remaining question is how the MAS should use recurring, sample-specific evidence to authorize bounded changes that persist across future tasks without destabilizing the system.

Evolving an MAS raises three questions about where to revise the system, when to revise it, and how to control regressions across revisions.
Locating a revision target requires a representation that exposes intermediate computation.
A wrong answer may result from missing computation, harmful information flow, or inadequate instructions.
Choosing when to revise requires distinguishing isolated failures from recurring patterns.
A revision should either improve accuracy or reduce token use without lowering accuracy.

We introduce \emph{EMAS}, an Evolving Multi-Agent System that consistently learns from new samples.
Building on GPTSwarm's graph view, EMAS represents an MAS as a graph of \emph{Steps} connected by directed edges.
Each Step performs one fine-grained LLM call, and the edges carry intermediate artifacts between Steps.
EMAS converts Step-level traces into structured diagnoses that specify a revision operation and target.
It generates a candidate revision only when the same structured diagnosis recurs across tasks.
The candidate revision is applied only if paired Validation shows that it outperforms the current MAS.
EMAS supports two directions of evolution, improving accuracy and reducing token use while maintaining accuracy.
The validation set is balanced, containing equal numbers of correct and incorrect samples to promote stable evolution.

\begin{figure}[t]
    \centering
    \includegraphics[width=\textwidth]{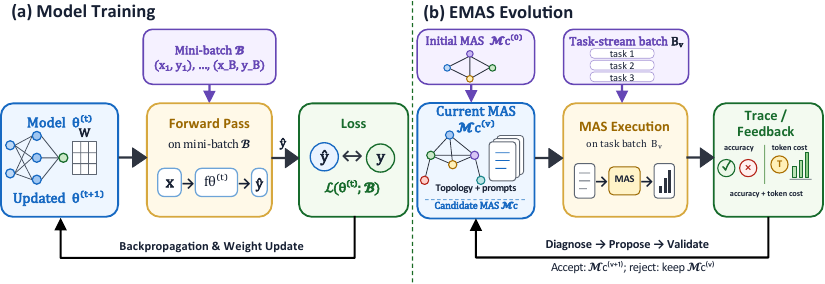}
    \caption{\textbf{Model training updates numerical weights, while EMAS evolves the executable system around a frozen language model.}
    \textbf{(a)} A mini-batch loss drives $\theta^{(t)}\!\rightarrow\theta^{(t+1)}$ through backpropagation.
    \textbf{(b)} The current category-specific MAS $\mathcal{M}_c^{(v)}$ produces traces and accuracy--cost feedback on task-stream executions through its topology and prompts.
    Evidence accumulated across tasks supports a discrete candidate $\widetilde{\mathcal{M}}_c$.
    The candidate becomes $\mathcal{M}_c^{(v+1)}$ only when paired Validation against the current MAS meets the objective-specific acceptance criterion.
    Otherwise, $\mathcal{M}_c^{(v)}$ is retained.}
    \vspace{-0.3cm}
    \label{fig:emas-motivation}
\end{figure}

Figure~\ref{fig:emas-motivation} contrasts parameter learning with system-level learning in EMAS.
Both use experience to update reusable state that shapes future behavior.
Model training updates numerical weights, whereas EMAS keeps the LLM fixed and updates system components.
An accepted revision becomes persistent MAS state and is used on subsequent tasks.

We evaluate EMAS on four benchmarks with two frozen LLM backbones.
EMAS attains the highest task-weighted overall accuracy for both backbones and is best or tied in six of eight model--benchmark settings.
Within two evolution epochs, EMAS achieves relative gains of 6.30\% and 20.10\% in task-weighted accuracy on Kimi-K2-6 and Qwen3.6-27B, respectively.
Extending Game24 evolution to Epoch 15 allows EMAS to reach 96.45\% accuracy on Kimi-K2-6, compared with 95.34\% within the first two epochs, while also reducing tokens per task.
On Qwen3.6-27B, EMAS reaches 94.73\% accuracy, compared with 79.90\% within the first two epochs, while reducing tokens per task by 46.77\%.
Among accepted revisions, successful accuracy-directed revisions improve accuracy by 4.46 percentage points on average, while regressive revisions reduce it by 3.03 points on average.
Together, these results show overall progress despite local regressions.

Controlled ablations further demonstrate the importance of recurrence and Validation for stable evolution.
On Game24 with Qwen3.6-27B, relative to Full EMAS, a committed transition is 1.14 times as likely to regress under the single-trace variant and 1.57 times as likely without the Validation gate.
The average accuracy loss per regressive transition is also 2.18 times as large under the single-trace variant and 18.05 times as large without the Validation gate.

Our contributions are threefold.
\begin{itemize}
    \item We formulate system-level evolution for MASs, where experience from subsequent tasks updates prompts and topology while the LLM weights remain fixed.
    \item We introduce EMAS, which turns recurring task experience into validated system updates for improving accuracy and reducing token use.
    \item Across four benchmarks and two frozen LLMs, we show that EMAS improves accuracy and can reduce token use, while recurrence and Validation make evolution more stable.
\end{itemize}

\section{Related Work}
\label{sec:related-work}

EMAS draws on automated agent design, execution-grounded optimization, and system self-improvement.
Prior work shows that agentic programs can be represented and searched, execution evidence can guide revisions, and candidate updates can be evaluated empirically.
EMAS focuses on the evolution of MAS on both accuracy and token usage.
We organize related work around three decisions in an evolution.
These decisions concern which reusable system state can change, which execution evidence justifies a revision proposal, and which comparison justifies committing the proposal as the next Version.

\subsection{Agentic Programs as Reusable System State}
\label{sec:related-work:learnable-design}

MAS is the orchestration of LLM, this allows the system to learn with fixed language model parameters.
AgentOptimizer treats callable functions as learnable parameters and revises them using execution histories~\citep{zhang2024agentoptimizer}.
Symbolic Learning updates prompts, tools, and pipeline structure using trajectory-derived language gradients for parameter updates and atomic edit operations for pipeline updates~\citep{ou2025symboliclearning}.
GPTSwarm makes the external program explicit as a computational graph and optimizes both node prompts and graph connectivity~\citep{zhuge2024gptswarm}.
Automated design methods such as ADAS and AFlow search over complete agent programs and code-represented workflows~\citep{hu2025adas,zhang2025aflow}.
MASS optimizes prompts and topology, and EvoFlow evolves workflow populations for task-adaptive reuse~\citep{zhou2026mass,zhang2025evoflow}.
Together, these methods establish prompts, computation, information flow, and complete agent programs as mutable system state above frozen model weights.

\subsection{From Execution Traces to Revision-Worthy Evidence}
\label{sec:related-work:execution-revision}

The final result indicates success or failure while system level revision requires fine-grained execution evidence.
Trace and its OptoPrime optimizer use execution traces and output feedback to revise heterogeneous program parameters, including prompts and code~\citep{cheng2024trace}.
Semantic Backpropagation propagates natural-language feedback through an agentic graph, aggregates it across queries, and admits parameter updates through a validation gate~\citep{wang2024semanticbackprop}.
Automated failure attribution shows the importance and difficulty of identifying the agent and step responsible for a multi-agent failure~\citep{zhang2025failureattribution}.
Reflexion converts task feedback into verbal reflections, and ExpeL extracts reusable insights from experience across tasks~\citep{shinn2023reflexion,zhao2024expel}.
These methods establish the value of internal execution evidence.
The same incorrect answer can have several causes, including missing computation, harmful information flow, and inadequate instruction.

Trajectory-driven methods already use richer execution evidence to construct and test revisions.
GEPA reflects on reasoning, tool calls, and outputs to propose and evaluate prompt updates, then combines complementary lessons along a Pareto frontier~\citep{agrawal2026gepa}.
MASPO mines historical multi-agent traces for local--global misalignment and jointly evolves role prompts within a fixed topology~\citep{wang2026maspo}.
These methods support trajectory-conditioned revision, cross-query aggregation, and candidate testing.
EMAS uses recurring diagnosis patterns across execution traces as evidence for revision.

\subsection{From Candidate Evaluation to Persistent Version Transitions}
\label{sec:related-work:persistent-improvement}

Prior methods use empirical evaluation to screen candidate changes.
Semantic Backpropagation uses a validation gate, Symbolic Learning supports rollback after evaluation, and GEPA tests alternative prompts~\citep{wang2024semanticbackprop,ou2025symboliclearning,agrawal2026gepa}.
OPRO generates new prompts from previously evaluated candidates, and Promptbreeder evolves prompt populations using training-set fitness~\citep{yang2024opro,fernando2024promptbreeder}.

Persistent change also appears in recursive and self-evolving systems.
The Darwin G\"odel Machine modifies agent code and uses empirical evaluation and an open-ended archive to guide further self-improvement~\citep{zhang2026dgm}.
Symbolic Learning similarly frames updates to deployed agentic programs as self-evolution~\citep{ou2025symboliclearning}.
EMAS studies stable MAS evolution driven by recurring execution evidence and empirical evaluation.

\begin{figure}[t]
    \centering
    \includegraphics[width=\textwidth]{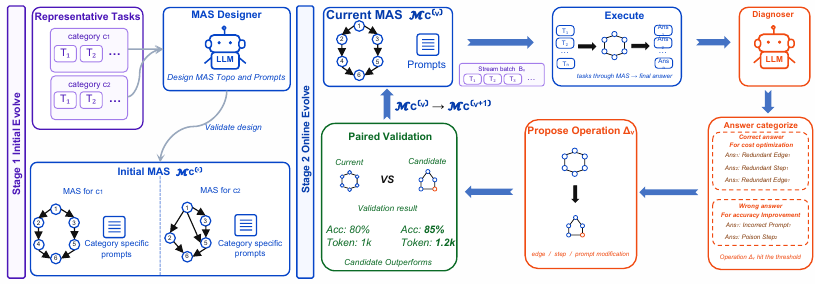}
    \caption{EMAS Evolution Pipeline Overview.}
    \label{fig:emas-methodology}
\end{figure}

\section{Methodology}
\label{sec:method}

\subsection{Problem Formulation and EMAS Overview}
\label{sec:method:overview}

EMAS evolves each MAS either to improve accuracy or to reduce cost while preserving accuracy.
To support more effective diagnosis, EMAS represents each MAS as a graph in which nodes correspond to LLM calls and edges transfer artifacts.
EMAS further increases the graph's granularity by defining each node as an atomic operation, thereby decomposing an Agent's responsibility across multiple nodes.
To enable targeted optimization for different classes of tasks, EMAS constructs and evolves a separate MAS for each task category.
For each category $c$, the current MAS at version $v$ is
\begin{equation}
    \mathcal{M}_{c}^{(v)}
    = \left(G_{c}^{(v)}, P_{c}^{(v)}\right),
    \label{eq:effective-mas-state}
\end{equation}
where $G_c^{(v)}$ specifies the graph's LLM-call nodes and edges, and $P_c^{(v)}$ specifies the prompt for each node.
A version therefore denotes a particular MAS state.
Only an accepted revision advances $v$; a rejected candidate leaves the current version unchanged.
EMAS supports two evolution objectives: repairing accuracy failures and reducing execution token cost without decreasing aggregate Validation accuracy.
The specific data construction and evolution configuration are reported in Table~\ref{tab:eval-setup} and Appendix~\ref{sec:appendix-results}.

The evolution lifecycle contains two checkpoints.
Step-level traces first produce localized, structured revision hypotheses.
EMAS generates a candidate only when the same hypothesis recurs across enough distinct Evolve samples.
This requirement ensures that only recurring patterns are considered and prevents outlier failures from triggering revisions.
The candidate is then compared with the current version on the same Validation set.
This comparison on the Validation set determines whether the candidate revision is accepted.

Figure~\ref{fig:emas-methodology} summarizes this process.
Initial Evolve constructs the initial MAS $\mathcal{M}_{c}^{(0)}$.
Online Evolve repeatedly executes samples, diagnoses traces, groups failure patterns, proposes a new revision after the corresponding support threshold is reached, and commits the candidate only when paired Validation satisfies the corresponding acceptance rule.

\subsection{Executable and Editable MAS State}
\label{sec:method:system-layer}

An \emph{Agent} is a role-bearing component responsible for multiple internal tasks.
EMAS exposes these tasks as fine-grained \emph{Steps}, which serve as the atomic units of execution.
Each Step performs one semantically atomic task through a single LLM call.
An Agent may therefore comprise one or more Steps, and the Step graph makes its internal computation and communication observable.

The graph $G_c^{(v)}=(V_c^{(v)},E_c^{(v)})$ contains LLM-call Steps $s\in V_c^{(v)}$ and directed artifact-flow Edges $E_c^{(v)}$.
The prompt state $P_c^{(v)}$ comprises system-, category-, phase-, and step-level instructions, enabling fine-grained control over execution.
In topological order, Step $s$ receives the original task, its prompt, and the artifacts of its direct predecessors.
The resulting MAS is therefore represented as a directed acyclic graph.
This structure makes it easier to monitor and diagnose the system.
Steps specify computation, Edges specify information flow, and prompts specify instructions.

For a benchmark $b$ with category set $\mathcal{C}_b$, the maintained MAS state is the collection of its category-specific systems:
\begin{equation}
    \mathcal{M}_{b}^{(\mathbf{v})}
    = \left\{\mathcal{M}_{c}^{(v_c)} \mid c\in\mathcal{C}_b\right\},
    \qquad
    \mathbf{v}=(v_c)_{c\in\mathcal{C}_b}.
    \label{eq:benchmark-mas-collection}
\end{equation}
To improve the precision of topology and prompt evolution, EMAS evolves one MAS for each task category within a benchmark.

\subsection{Initializing the MAS State}
\label{sec:method:bootstrap}

EMAS begins evolution from an initial MAS.
During Initial Evolve, EMAS deterministically selects a small set of representative inputs for each category.
The category Designer jointly processes these representative inputs and produces a complete design containing semantically atomic Steps, Edges, and fine-grained prompts.

Structural validation and repair enforce valid graph references, a valid topology, and complete prompt coverage.
Once validated, the design is stored as the initial MAS $\mathcal{M}_c^{(0)}$.
Initial Evolve leverages the design capabilities of LLMs to construct the initial MAS as the LLM can design very well.
Prior work similarly uses LLMs to automatically design agentic systems and generate agent workflows~\citep{hu2025adas,zhang2025aflow}.
Initial Evolve therefore supplies a valid but potentially imperfect initial state.
Online Evolve starts from this state, and its executions expose recurring limitations that can support subsequent revisions.

\subsection{Online Evolve: From Step Traces to Operational Revision Hypotheses}
\label{sec:method:experience}

For sample $i$, Online Evolve executes the current $\mathcal{M}_c^{(v)}$ and records an observation $\tau_i^{(v)}$ containing the diagnosis derived from the trace.
Because the observation preserves canonical Step, Edge, Phase, and prompt locations, a system-level outcome can be translated into a localized and operationally testable revision hypothesis.
Incorrect outputs produce accuracy-oriented hypotheses, whereas correct outputs produce cost-oriented hypotheses.
Accuracy diagnosis searches for a recoverable defect that affects correctness.
Cost diagnosis searches for avoidable computation, information transfer, or instructions while preserving the correct outcome.

A trace can yield multiple records, each canonicalized as $h=(o,d,a,\ell)$.
Here, $o\in\{\textsc{accuracy},\textsc{cost}\}$ denotes the objective, $d$ denotes the defect class, $a$ denotes the associated operation, and $\ell$ denotes its canonical location.

EMAS can modify three components of a MAS: Steps, Edges, and prompts.
For Steps, \textsc{add-node} addresses a missing operation, \textsc{remove-node} addresses a redundant or harmful operation, and \textsc{split-node} addresses a non-atomic operation.
For Edges, \textsc{add-edge} supplies missing information to a Step, whereas \textsc{remove-edge} eliminates redundant or harmful artifact flow.
For prompts, \textsc{prompt-only} revises the relevant prompt to provide better instructions.

For example, if a Verifier Step does not receive a Solver artifact, the Diagnoser records a missing-information, \textsc{add-edge} hypothesis at the canonical Solver-to-Verifier location.
Such a hypothesis can support candidate generation when it recurs across multiple samples.

\subsection{Recurrence-Gated Candidate Construction}
\label{sec:method:evidence-gated-revisions}

Stable evolution requires ignoring outlier failures and revising only in response to failure patterns that recur across multiple samples.
EMAS therefore accumulates support from distinct samples before constructing a candidate.
For each hypothesis $h$, EMAS maintains an evidence buffer containing the distinct samples whose traces support $h$.
Candidate construction is triggered only when the number of supporting samples reaches the operation-specific threshold $\kappa_a$.

Operation-specific thresholds reflect differences in revision risk.
Adding a Step or Edge is generally less risky because it typically introduces redundancy rather than removing existing computation or information flow.
By contrast, removal is riskier because it can eliminate computation or information flow required by some samples.
The threshold settings appear in Appendix~\ref{sec:appendix-results}.

Each candidate is restricted to one primary change at $\ell$ and a narrowly scoped closure of directly affected Edges and prompts.
\textsc{add-node} adds exactly one Step and may add only Edges incident to the new Step and input-aligned prompts, whereas \textsc{remove-node} removes exactly one Step and may update only incident or bypass Edges and input-aligned prompts.
\textsc{split-node} preserves the target Step, adds same-Phase Steps, and may update only split-group Edges and prompts together with direct-successor prompts.
\textsc{add-edge} and \textsc{remove-edge} change exactly one Edge and may revise only the target Step's prompt.
\textsc{prompt-only} revises exactly the slot encoded by $\ell$, leaving topology and all other prompts unchanged.

\subsection{Paired Validation and Persistent Version Transition}
\label{sec:method:validation-transitions}

A structurally valid and evidence-aligned candidate may still fail to improve the system.
The candidate and current MAS are therefore evaluated on the same fixed Validation set $\mathcal{D}_c^{\mathrm{val}}$.
Both systems use the same execution and evaluation configuration to ensure a fair comparison.
For a system $M$, define its aggregate correct count as \(C(M)=\sum_{i\in\mathcal{D}_c^{\mathrm{val}}}r_i(M)\) and its total execution token cost as \(T(M)=\sum_{i\in\mathcal{D}_c^{\mathrm{val}}}\operatorname{tokens}_i(M)\).
EMAS applies the following objective-specific acceptance rules:
\begin{equation}
\begin{aligned}
    \operatorname{Accept}_{\mathrm{acc}}(\widetilde{\mathcal{M}}_c)
    &\iff C(\widetilde{\mathcal{M}}_c)>C(\mathcal{M}_c^{(v)}),\\
    \operatorname{Accept}_{\mathrm{cost}}(\widetilde{\mathcal{M}}_c)
    &\iff C(\widetilde{\mathcal{M}}_c)\geq C(\mathcal{M}_c^{(v)})
    \ \land\ 
    T(\widetilde{\mathcal{M}}_c)<T(\mathcal{M}_c^{(v)}).
\end{aligned}
    \label{eq:acceptance-rules}
\end{equation}
An accuracy-oriented candidate is accepted only if it strictly increases the aggregate correct count.
A cost-oriented candidate is accepted only if it maintains or increases the aggregate correct count while strictly reducing the total execution token cost.
These empirical Validation rules do not guarantee improvement on held-out Test tasks.

On rejection, EMAS retains the current $\mathcal{M}_c^{(v)}$ and records the rejected candidate.
On acceptance, EMAS promotes the candidate to $\mathcal{M}_c^{(v+1)}$ and makes it the current MAS.
In either outcome, EMAS consumes every buffered diagnosis record associated with a sample that supported the processed trigger.
Following acceptance, EMAS re-executes the remaining active samples under $\mathcal{M}_c^{(v+1)}$ to obtain up-to-date diagnoses.
This replay prevents stale evidence from triggering subsequent candidate proposals.

\section{Experiments}
\label{sec:experiments}

We evaluate EMAS along three dimensions: whether experience-guided evolution improves MAS performance, how performance changes over the evolution, and whether recurrence and paired Validation stabilize the evolution process.
We use two frozen LLM backbones, Qwen3.6-27B and Kimi-K2-6, on four benchmarks spanning mathematical reasoning, code generation, planning, and combinatorial search.
Table~\ref{tab:eval-setup} summarizes the data partitions and evaluation setup, with further details in Appendix~\ref{sec:appendix-results}.

\subsection{Overall Performance Across Backbones and Benchmarks}
\label{sec:experiments:main-results}

Table~\ref{tab:eval-main} compares Genesis, the Initial MAS produced by Initial Evolve, and the evolved MAS produced after two epochs of Online Evolve.
It also includes Genesis+SOP, AFlow, and ADAS as comparison methods.
Genesis uses only a generic problem-solving instruction and task-specific output constraints when required.
During Initial Evolve, the Designer generates a standard operating procedure (SOP) and decomposes it into an executable MAS, reported as Initial MAS.
The Genesis+SOP baseline adds the SOP to the Genesis prompt but does not decompose it into an MAS.
Its lower task-weighted accuracy than Initial MAS with both backbones indicates that the executable MAS structure contributes beyond the SOP text alone.
After two epochs of Online Evolve, EMAS achieves higher task-weighted accuracy than AFlow and ADAS with both backbones.

\begin{table}[!htbp]
  \centering
  \caption{\textbf{Main results across two models and four benchmarks under a fixed two-epoch evolution budget.} Relative to the Initial MAS, EMAS improves accuracy on seven of eight model--benchmark pairs and ties the remaining pair, raising task-weighted overall accuracy from 90.11\% to 95.79\% with Kimi-K2-6 and from 73.42\% to 88.18\% with Qwen3.6-27B. Token changes are mixed across settings. Each cell reports mean accuracy (\%; $\uparrow$) / mean tokens per task (thousands; $\downarrow$); Overall is task-weighted across benchmarks. All competing baselines average repeats 1--3. Genesis, Initial MAS, and EMAS use three runs. }
  \label{tab:eval-main}
  \begingroup
  \EvalTableMainStyle
  \begin{tabular*}{\textwidth}{@{\extracolsep{\fill}}lccccc@{}}
    \toprule
    & \multicolumn{5}{c}{\textbf{Accuracy (\%) $\uparrow$ / Tokens per task (k) $\downarrow$}} \\
    \cmidrule(lr){2-6}
    \textbf{Method} & \textbf{Overall} & \textbf{Math} & \textbf{MBPP} & \textbf{PlanBench} & \textbf{Game24} \\
    \midrule
    \rowcolor{TableGroup} \multicolumn{6}{l}{\textbf{Kimi-K2-6}} \\
    Genesis & 54.92 / 0.83 & 78.37 / 0.61 & 68.91 / 0.28 & 14.40 / 2.17 & 15.69 / 0.12 \\
    Initial MAS & 90.11 / 14.81 & 97.43 / 14.19 & \textbf{95.85 / 9.53} & 83.51 / 23.34 & 68.38 / 8.84 \\
    Genesis+SOP & 35.88 / 1.64 & 39.10 / 1.48 & 85.66 / 1.35 & 14.75 / 2.39 & 18.38 / 1.33 \\
    AFlow & 84.43 / 6.77 & 89.80 / 8.12 & 89.81 / 3.69 & 60.03 / 7.62 & 95.10 / 2.78 \\
    ADAS & 67.39 / 3.61 & 81.00 / 1.98 & 82.73 / 4.12 & 3.49 / 5.26 & \textbf{96.20 / 6.93} \\
    \rowcolor{TablePrimary} EMAS & \textbf{95.79 / 17.07} & \textbf{98.83 / 13.84} & \textbf{95.85 / 9.53} & \textbf{88.13 / 32.27} & 95.34 / 12.97 \\
    \addlinespace[2pt]
    \rowcolor{TableGroup} \multicolumn{6}{l}{\textbf{Qwen3.6-27B}} \\
    Genesis & 31.98 / 0.30 & 38.33 / 0.13 & 58.38 / 0.27 & 14.14 / 0.90 & 14.95 / 0.12 \\
    Initial MAS & 73.42 / 7.44 & 97.50 / 8.22 & 55.09 / 5.16 & 35.08 / 9.54 & 51.72 / 3.24 \\
    Genesis+SOP & 26.84 / 1.04 & 28.87 / 0.87 & 59.93 / 0.94 & 13.35 / 1.73 & 14.83 / 0.77 \\
    AFlow & 75.53 / 10.40 & 97.60 / 4.63 & \textbf{93.09 / 4.52} & 51.66 / 33.65 & 15.44 / 3.16 \\
    ADAS & 74.81 / 1.86 & 90.57 / 1.49 & 86.70 / 0.73 & 33.07 / 3.68 & 67.03 / 1.46 \\
    \rowcolor{TablePrimary} EMAS & \textbf{88.18 / 8.01} & \textbf{97.70 / 7.03} & 89.12 / 1.95 & \textbf{68.67 / 14.71} & \textbf{79.90 / 6.54} \\
    \bottomrule
  \end{tabular*}
  \endgroup
\end{table}

Evolution improves strong initial systems across models and tasks.
EMAS obtains the highest task-weighted accuracy with both backbones and is best in six of the eight model--benchmark settings.
Relative to the Initial MAS, accuracy improves in seven settings.
Overall accuracy rises from 90.11\% to 95.79\% for Kimi-K2-6 and from 73.42\% to 88.18\% for Qwen3.6-27B, corresponding to relative gains of 6.30\% and 20.10\%.
These gains across both backbones and all four task families show that evolution improves MAS performance across models and task types.
Appendix Table~\ref{tab:eval-genesis-best-composite} reports the complete three-run results and confidence intervals.

Evolution gains depend on the headroom and maturity of the MAS.
Qwen starts from a weaker Initial MAS and realizes a larger relative gain, with its largest benchmark improvements on MBPP, PlanBench, and Game24.
By contrast, Kimi starts from a stronger Initial MAS, and its near-saturated Math setting leaves less room for accuracy improvement.
This pattern suggests that recurring failures mainly drive accuracy repair when substantial headroom remains, whereas systems closer to saturation provide more opportunities for cost-directed revision.
Consistent with this interpretation, Appendix Table~\ref{tab:eval-headroom-alignment} shows that successful accuracy revisions yield larger gains when their predecessor has lower accuracy.

EMAS can improve accuracy and reduce token use in the same setting.
On Qwen--MBPP, EMAS raises accuracy from 55.09\% to 89.12\% while reducing tokens per task from 5.16k to 1.95k, a 62.2\% reduction.
Under the two-epoch budget, however, evolution primarily improves accuracy, and not every setting achieves a token reduction.
Two epochs may therefore be insufficient to realize both accuracy and cost improvements across all settings.
Some categories retain the Initial MAS as their best checkpoint, whereas others reach their best state only after several committed revisions (Appendix Table~\ref{tab:eval-best-version-detail}).

\subsection{Evolution Dynamics Across Versions}
\label{sec:experiments:evolution-dynamics}

\begin{figure}[t]
  \centering
  \includegraphics[width=\textwidth]{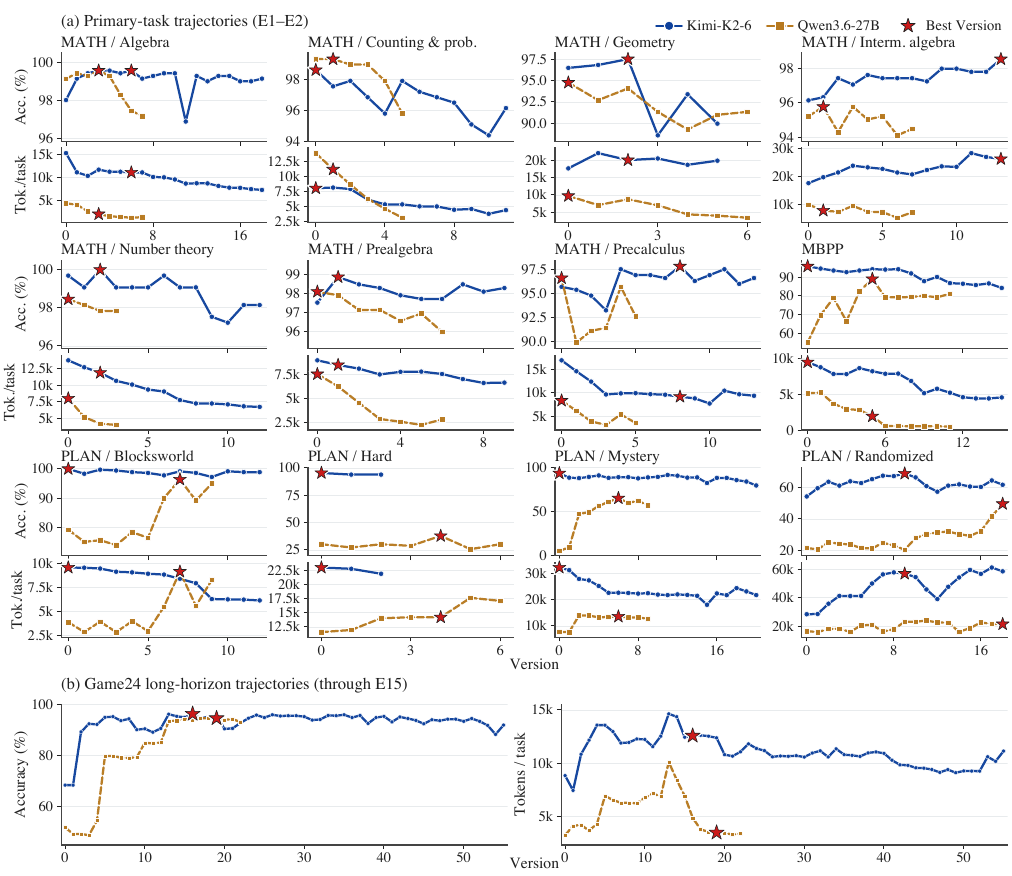}
  \input{assets/figures/eval_version_small_multiples_caption.tex}
  \vspace{-0.3cm}
  \label{fig:eval-version-small-multiples}
\end{figure}

Figure~\ref{fig:eval-version-small-multiples} shows the evolution curves for all configurations, together with the extended evolution on Game24.
Overall, accuracy increases and token use decreases.
This trend is clearest in the extended evolution, where Qwen achieves high accuracy while keeping its token cost close to that of the Initial MAS.

Evolution changes character as a system matures.
Systems with substantial initial headroom, most visibly Qwen on MBPP, PlanBench, and Game24, exhibit large accuracy jumps.
In high-accuracy Math categories, accuracy moves within a narrower range while token cost can continue to fall.
Across trajectories, accuracy-directed evolution is more prominent in the earlier stages, whereas cost-directed evolution becomes more prominent once correctness approaches saturation.
Multiple categories contain several accuracy--token Pareto-optimal Versions (Appendix Figure~\ref{fig:appendix-version-pareto}), showing that evolution constructs a family of useful operating points.

Individual revisions can move accuracy or token use in the wrong direction on the test set, after which later Versions may recover and surpass earlier states.
This behavior has two causes.
First, EMAS uses the Validation set to accept modifications to the MAS, but acceptance on Validation does not guarantee improvement on the Test set.
Second, EMAS uses vLLM as the backend for a multi-node MAS, so small numerical instabilities can occasionally produce different answers.
EMAS therefore uses a checkpoint mechanism to retain the best-performing Version.

Extended evolution reveals backbone-dependent evolution rates.
Extending Game24 to Epoch 15 improves the best Kimi checkpoint from 95.34\% accuracy and 12,971 tokens per task to 96.45\% and 12,585 tokens, while Qwen improves from 79.90\% and 6,539 tokens to 94.73\% and 3,481 tokens (Appendix Table~\ref{tab:eval-game24-passes}).
Kimi enters the high-accuracy regime earlier and subsequently makes smaller accuracy--cost refinements; Qwen requires a longer repair phase but realizes a larger eventual gain.
Thus, the same evolution procedure adapts to the capability of each backbone and can continue to expand the reachable accuracy--cost frontier beyond the main two-epoch budget.
Together, these trajectories demonstrate persistent evolution.

\subsection{Ablation of Evidence Accumulation and Validation Gating}
\label{sec:experiments:ablations}

\begin{wraptable}{r}{0.53\textwidth}
  \vspace{-2\baselineskip}
  \centering
  \caption{\textbf{Evidence accumulation and Validation gating stabilize EMAS evolution.} On Game24 with Qwen3.6-27B, we compare Full EMAS, $\kappa=1$, and removal of the Validation gate from V0 to V20. Full EMAS achieves the highest accuracy, lowest token use, and smallest average per-transition regression. Results are averaged over three runs.}
  \label{tab:eval-core-ablations}
  \begingroup
  \EvalTableStandardStyle
  \begin{tabular*}{\linewidth}{@{\extracolsep{\fill}}lrrr@{}}
    \toprule
    Ablation & Acc. (\%) $\uparrow$ & Tokens (k) $\downarrow$ & Avg. reg. (pp) $\downarrow$ \\
    \midrule
    Full EMAS & \textbf{94.73} & \textbf{3.48} & \textbf{0.28} \\
    $\kappa=1$ & 89.58 & 12.07 & 0.70 \\
    No gate & 65.56 & 4.26 & 7.93 \\
    \bottomrule
  \end{tabular*}
  \endgroup
  \vspace{-1\baselineskip}
\end{wraptable}

We use an ablation study to examine the two components that control stable Version transitions on Game24 with Qwen3.6-27B.
Setting $\kappa=1$ removes evidence accumulation, allowing a single diagnosis to trigger a revision.
Removing the Validation gate allows generated revisions to enter the Version trajectory without paired screening.
The two ablations therefore test complementary decisions in EMAS: whether a diagnosed defect has sufficient recurring support and whether the proposed repair is safe to commit.

Recurrence limits the damage and complexity induced by noisy evidence.
Full EMAS and $\kappa=1$ have seven and eight regressive transitions, respectively, but their cumulative accuracy losses are 5.59 and 13.92 percentage points (Appendix Table~\ref{tab:eval-ablation-stability-detail}).
Thus, recurrence does not merely eliminate an occasional downward transition; it reduces the influence by idiosyncratic traces.

Validation distinguishes genuine efficiency from destructive simplification.
Without the gate, token use remains only moderately higher than that of Full EMAS (4.26k versus 3.48k), but the best accuracy falls from 94.73\% to 65.56\%, and the average per-transition regression rises from 0.28 to 7.93 percentage points.
Paired Validation prevents such destructive simplification, as well as faulty accuracy-oriented repairs, from becoming persistent MAS state.

The two controls address different failure modes.
Recurrence improves the evidence used to identify a system-level defect, whereas Validation tests the quality of the concrete candidate proposed for that defect.
The complete V0--V20 trajectories in Appendix Figure~\ref{fig:eval-ablation-complete} further show that this advantage persists across the evolution path rather than arising from a single isolated checkpoint.

\section{Conclusion and Limitations}
\label{sec:conclusion}

Many automated MAS design pipelines stop changing just as execution on new tasks begins to provide their most relevant evidence.
Yet making the MAS mutable creates the opposite risk because beneficial revisions persist, and mistaken ones do as well.
EMAS instead treats evolution as a revision-authorization problem by mapping traces to localized hypotheses, requiring recurrence across tasks, and committing a bounded topology or prompt revision only after paired Validation.
This turns task-local experience into auditable system state and shifts the objective from generating plausible edits to deciding which revisions may shape future behavior.
Within two evolution epochs, EMAS attains the highest task-weighted accuracy among the compared methods for both backbones and is best or tied in six of eight model--benchmark settings.
On Qwen--MBPP, its highest-accuracy observed checkpoint raises accuracy from 55.09\% to 89.12\% while reducing tokens per task by 62.2\%.
Across trajectories, systems with substantial headroom first make large accuracy repairs, whereas systems near saturation increasingly refine cost.
The ablations further show that recurrence and Validation limit damage from noisy evidence and destructive candidates.

Evolution nevertheless remains non-monotonic.
Repeated use of a fixed Validation set can induce adaptive selection, aggregate improvement need not preserve every solved task, and 39 of 93 committed E2 revisions with exact-predecessor comparisons regress on Test.
Moreover, the headline results use the best checkpoints observed retrospectively on Test.
These checkpoints show that better states are reachable, but they do not provide a deployment-time rule for identifying them.
Recurrence also does not imply causal equivalence because the current structured key can merge traces with distinct causes, while a single bounded edit may miss repairs that require coordinated changes.
EMAS also assumes correctness feedback, incurs substantial paired-Validation cost, and has been evaluated on only two backbones and four benchmarks.
Future work should combine rolling validation, retention-aware acceptance, coordinated revisions, and cheaper screening.
A frozen model need not imply a frozen executor, but its evolution must generalize, preserve prior capabilities, and pay for itself.

\clearpage

\subsection*{AI use statement}

Generative AI tools were used to assist with editing software code and
\LaTeX{}, creating and revising scientific figures and tables, organizing the
manuscript, improving readability, and checking formatting and references. All
AI-assisted code, text, and artifacts were reviewed by the authors. Numerical
results were checked against the underlying experimental records, generated
tables, and audit files. The authors retained final control over the scientific
claims and take responsibility for the complete content of this work, including
all AI-assisted material.

\subsection*{Reproducibility statement}

The method and its state-transition procedure are specified in
Section~\ref{sec:method}; the evaluation protocol, main comparisons, and
ablations are described in Section~\ref{sec:experiments} and the appendix. The
supplementary material includes the recorded result bundle, provenance
manifests and checksums, and the scripts used to regenerate and validate the
paper-facing tables and figures. These checks reject incomplete records,
inconsistent aggregates, unregistered artifacts, and failed numerical
invariants.

\clearpage

\bibliography{references/references}
\bibliographystyle{styles/iclr2027_conference}

\appendix
\clearpage
\raggedbottom

\section{Additional Experimental Results}
\label{sec:appendix-results}

\subsection{Experimental Data and Evaluation Scope}

Table~\ref{tab:eval-setup} summarizes the data used for evolution and
evaluation. Each benchmark is split into Train and Test. Test is fixed, shared
by both backbones, and held out during evolution.

For each backbone and category, Train is further divided into Evolve and
Validation based on the outcomes of V0 on the full Train split. Validation
includes both V0-correct and V0-incorrect tasks when possible. For categories
large enough, it contains at least 50 tasks. It can therefore be imbalanced when
V0 makes very few errors. For example, Kimi makes only three errors on the
eligible PlanBench blocks tasks, resulting in 47 correct and 3 incorrect tasks
in Validation. The remaining Train tasks are used for Evolve.

Evolve provides experience for system revision, Validation is used to accept
or reject revisions, and Test is used for final evaluation.

\begin{table}[!htbp]
  \centering
  \caption{\textbf{Data partitions for evolution and evaluation.} Each benchmark is split into 80\% Train and 20\% Test data. For each model and category, the Initial MAS (V0) is run on Train, which is then divided into Evolve and Validation. Validation includes both V0-correct and V0-incorrect tasks when possible, but can be imbalanced when one outcome is scarce and the minimum split size must be maintained. Validation cells report the split size followed by the corresponding correct/incorrect composition in parentheses. Test is shared across models and remains held out throughout evolution.}
  \label{tab:eval-setup}
  \begingroup
  \EvalTableDenseStyle
  \setbox0=\hbox{%
  \begin{tabular}{lllrrrr}
    \toprule
    \multirow{2}{*}{Benchmark} & \multirow{2}{*}{Model} & \multirow{2}{*}{Category} & \multicolumn{3}{c}{Train} & \multirow{2}{*}{Test} \\
    \cmidrule(lr){4-6}
    & & & Total & Evolve & Validation (V0 R/W) & \\
    \midrule
    Math & Kimi-K2-6 & algebra & 950 & 760 & 190 (95/95) & 237 \\
     &  & count./prob. & 379 & 329 & 50 (26/24) & 95 \\
     &  & geometry & 383 & 307 & 76 (38/38) & 96 \\
     &  & interm. algebra & 722 & 578 & 144 (72/72) & 181 \\
     &  & number theory & 432 & 382 & 50 (28/22) & 108 \\
     &  & prealgebra & 697 & 563 & 134 (67/67) & 174 \\
     &  & precalculus & 437 & 349 & 88 (44/44) & 109 \\
     &  & \textbf{Overall} & \textbf{4,000} & \textbf{3,268} & \textbf{732 (370/362)} & \textbf{1,000} \\
    \cmidrule(lr){2-7}
     & Qwen3.6-27B & algebra & 950 & 834 & 116 (58/58) & 237 \\
     &  & count./prob. & 379 & 329 & 50 (27/23) & 95 \\
     &  & geometry & 383 & 307 & 76 (38/38) & 96 \\
     &  & interm. algebra & 722 & 578 & 144 (72/72) & 181 \\
     &  & number theory & 432 & 382 & 50 (31/19) & 108 \\
     &  & prealgebra & 697 & 583 & 114 (57/57) & 174 \\
     &  & precalculus & 437 & 349 & 88 (44/44) & 109 \\
     &  & \textbf{Overall} & \textbf{4,000} & \textbf{3,362} & \textbf{638 (327/311)} & \textbf{1,000} \\
    \midrule
    MBPP & Kimi-K2-6 & all & 771 & 717 & 54 (27/27) & 193 \\
     &  & \textbf{Overall} & \textbf{771} & \textbf{717} & \textbf{54 (27/27)} & \textbf{193} \\
    \cmidrule(lr){2-7}
     & Qwen3.6-27B & all & 771 & 711 & 60 (30/30) & 193 \\
     &  & \textbf{Overall} & \textbf{771} & \textbf{711} & \textbf{60 (30/30)} & \textbf{193} \\
    \midrule
    PlanBench & Kimi-K2-6 & blocks. & 480 & 430 & 50 (47/3) & 120 \\
     &  & blocks. hard & 88 & 70 & 18 (12/6) & 22 \\
     &  & mystery & 480 & 418 & 62 (31/31) & 120 \\
     &  & randomized & 480 & 384 & 96 (48/48) & 120 \\
     &  & \textbf{Overall} & \textbf{1,528} & \textbf{1,302} & \textbf{226 (138/88)} & \textbf{382} \\
    \cmidrule(lr){2-7}
     & Qwen3.6-27B & blocks. & 480 & 384 & 96 (48/48) & 120 \\
     &  & blocks. hard & 88 & 70 & 18 (9/9) & 22 \\
     &  & mystery & 480 & 388 & 92 (46/46) & 120 \\
     &  & randomized & 480 & 384 & 96 (48/48) & 120 \\
     &  & \textbf{Overall} & \textbf{1,528} & \textbf{1,226} & \textbf{302 (151/151)} & \textbf{382} \\
    \midrule
    Game24 & Kimi-K2-6 & all & 1,090 & 872 & 218 (109/109) & 272 \\
     &  & \textbf{Overall} & \textbf{1,090} & \textbf{872} & \textbf{218 (109/109)} & \textbf{272} \\
    \cmidrule(lr){2-7}
     & Qwen3.6-27B & all & 1,090 & 872 & 218 (109/109) & 272 \\
     &  & \textbf{Overall} & \textbf{1,090} & \textbf{872} & \textbf{218 (109/109)} & \textbf{272} \\
    \bottomrule
  \end{tabular}}
  \noindent\makebox[\textwidth][c]{%
    \ifdim\wd0>\textwidth
      \resizebox{\textwidth}{!}{\box0}%
    \else
      \box0
    \fi}
  \endgroup
\end{table}

\subsection{Primary Two-Epoch Results and Checkpoint Provenance}

\begin{table}[H]
  \centering
  \caption{\textbf{EMAS improves mean accuracy over the Initial MAS in seven of eight model--benchmark settings and preserves it in the remaining setting.} Genesis, Initial MAS, and the category-wise EMAS results achieved within a fixed two-epoch evolution budget are compared with three external baselines on the same Test tasks. All competing baselines average repeats 1--3. Cells report mean accuracy and tokens per task with across-run 95\% confidence intervals.}
  \label{tab:eval-genesis-best-composite}
  \begingroup
  \EvalTableDenseStyle
  \begin{minipage}[t]{0.49\textwidth}
    \centering
    \vspace{0pt}%
    \EvalTablePanelTitle{Kimi-K2-6}
    \setbox0=\hbox{%
    \begin{tabular}{llcrr}
      \toprule
    Benchmark & Method & Runs & Accuracy [95\% CI] & Tokens/task [95\% CI] \\
      \midrule
    \multirow{6}{*}{Math} & Genesis & 3 & 78.37 [75.71, 81.02] & 609 [599, 619] \\
     & Initial MAS & 3 & 97.43 [96.31, 98.55] & 14186 [14108, 14264] \\
     & Genesis+SOP & 3 & 39.10 [38.77, 39.43] & 1485 [1476, 1493] \\
     & AFlow & 3 & 89.80 [89.22, 90.38] & 8121 [8105, 8137] \\
     & ADAS & 3 & 81.00 [80.42, 81.58] & 1983 [1974, 1993] \\
     & EMAS & 3 & 98.83 [98.45, 99.21] & 13842 [13785, 13898] \\
    \cmidrule(lr){1-5}
    \multirow{6}{*}{MBPP} & Genesis & 3 & 68.91 [67.62, 70.20] & 279 [230, 328] \\
     & Initial MAS & 3 & 95.85 [93.28, 98.43] & 9532 [9228, 9836] \\
     & Genesis+SOP & 3 & 85.66 [84.93, 86.40] & 1355 [1353, 1356] \\
     & AFlow & 3 & 89.81 [88.61, 91.01] & 3685 [3645, 3725] \\
     & ADAS & 3 & 82.73 [80.79, 84.66] & 4115 [4089, 4141] \\
     & EMAS & 3 & 95.85 [93.28, 98.43] & 9532 [9228, 9836] \\
    \cmidrule(lr){1-5}
    \multirow{6}{*}{PlanBench} & Genesis & 3 & 14.40 [13.75, 15.05] & 2171 [2094, 2247] \\
     & Initial MAS & 3 & 83.51 [81.16, 85.85] & 23345 [23048, 23642] \\
     & Genesis+SOP & 3 & 14.75 [14.24, 15.25] & 2392 [2367, 2416] \\
     & AFlow & 3 & 60.03 [59.12, 60.95] & 7618 [7594, 7642] \\
     & ADAS & 3 & 3.49 [2.57, 4.41] & 5260 [5239, 5281] \\
     & EMAS & 3 & 88.13 [86.78, 89.49] & 32265 [31699, 32831] \\
    \cmidrule(lr){1-5}
    \multirow{6}{*}{Game24} & Genesis & 3 & 15.69 [14.29, 17.08] & 119 [97, 141] \\
     & Initial MAS & 3 & 68.38 [62.39, 74.37] & 8838 [8764, 8912] \\
     & Genesis+SOP & 3 & 18.38 [18.04, 18.72] & 1335 [1327, 1343] \\
     & AFlow & 3 & 95.10 [94.58, 95.62] & 2782 [2611, 2952] \\
     & ADAS & 3 & 96.20 [95.68, 96.72] & 6927 [6878, 6977] \\
     & EMAS & 3 & 95.34 [91.65, 99.03] & 12971 [12165, 13776] \\
      \bottomrule
    \end{tabular}}
    \noindent\makebox[\linewidth][c]{%
      \ifdim\wd0>\linewidth
        \resizebox{\linewidth}{!}{\box0}%
      \else
        \box0
      \fi}
  \end{minipage}%
  \hfill
  \begin{minipage}[t]{0.49\textwidth}
    \centering
    \vspace{0pt}%
    \EvalTablePanelTitle{Qwen3.6-27B}
    \setbox0=\hbox{%
    \begin{tabular}{llcrr}
      \toprule
    Benchmark & Method & Runs & Accuracy [95\% CI] & Tokens/task [95\% CI] \\
      \midrule
    \multirow{6}{*}{Math} & Genesis & 3 & 38.33 [38.19, 38.48] & 126 [126, 126] \\
     & Initial MAS & 3 & 97.50 [96.84, 98.16] & 8224 [8204, 8244] \\
     & Genesis+SOP & 3 & 28.87 [28.54, 29.19] & 873 [873, 873] \\
     & AFlow & 3 & 97.60 [97.51, 97.69] & 4629 [4601, 4658] \\
     & ADAS & 3 & 90.57 [90.43, 90.71] & 1494 [1480, 1508] \\
     & EMAS & 3 & 97.70 [97.04, 98.36] & 7025 [6925, 7124] \\
    \cmidrule(lr){1-5}
    \multirow{6}{*}{MBPP} & Genesis & 3 & 58.38 [57.63, 59.12] & 274 [266, 282] \\
     & Initial MAS & 3 & 55.09 [49.74, 60.45] & 5155 [4811, 5498] \\
     & Genesis+SOP & 3 & 59.93 [59.65, 60.21] & 945 [943, 947] \\
     & AFlow & 3 & 93.09 [91.89, 94.30] & 4524 [4485, 4562] \\
     & ADAS & 3 & 86.70 [85.97, 87.43] & 730 [711, 749] \\
     & EMAS & 3 & 89.12 [87.83, 90.41] & 1949 [1857, 2041] \\
    \cmidrule(lr){1-5}
    \multirow{6}{*}{PlanBench} & Genesis & 3 & 14.14 [14.14, 14.14] & 899 [859, 940] \\
     & Initial MAS & 3 & 35.08 [32.24, 37.91] & 9536 [9336, 9736] \\
     & Genesis+SOP & 3 & 13.35 [13.35, 13.35] & 1727 [1721, 1732] \\
     & AFlow & 3 & 51.66 [50.42, 52.90] & 33646 [33247, 34045] \\
     & ADAS & 3 & 33.07 [32.16, 33.99] & 3683 [3676, 3691] \\
     & EMAS & 3 & 68.67 [66.58, 70.76] & 14713 [14672, 14754] \\
    \cmidrule(lr){1-5}
    \multirow{6}{*}{Game24} & Genesis & 3 & 14.95 [13.90, 16.01] & 123 [123, 123] \\
     & Initial MAS & 3 & 51.72 [48.26, 55.17] & 3238 [3076, 3399] \\
     & Genesis+SOP & 3 & 14.83 [14.63, 15.02] & 775 [774, 775] \\
     & AFlow & 3 & 15.44 [15.10, 15.78] & 3155 [3131, 3179] \\
     & ADAS & 3 & 67.03 [65.66, 68.41] & 1457 [1440, 1474] \\
     & EMAS & 3 & 79.90 [75.59, 84.22] & 6539 [6370, 6708] \\
      \bottomrule
    \end{tabular}}
    \noindent\makebox[\linewidth][c]{%
      \ifdim\wd0>\linewidth
        \resizebox{\linewidth}{!}{\box0}%
      \else
        \box0
      \fi}
  \end{minipage}%
  \endgroup
\end{table}

\begin{table}[H]
  \centering
  \caption{\textbf{Category-level provenance of the reported EMAS results.} For each model and category, the table identifies the Version at which the reported result was achieved within the fixed two-epoch evolution budget. When multiple Versions attain the same reported accuracy, ties favor fewer tokens and then the earlier Version. Version numbers index committed revisions within each category; V0 denotes the Initial MAS. Values are averaged over three runs.}
  \label{tab:eval-best-version-detail}
  \begingroup
  \EvalTableStandardStyle
  \begin{minipage}[t]{0.49\textwidth}
    \centering
    \vspace{0pt}%
    \EvalTablePanelTitle{Kimi-K2-6}
    \setbox0=\hbox{%
    \begin{tabular}{llrrr}
      \toprule
    Benchmark & Category & Reported Version & Accuracy (\%) & Tokens/task \\
      \midrule
    \multirow{7}{*}{Math} & algebra & V6 & 99.58 & 11,153 \\
     & count./prob. & V0 & 98.60 & 8,095 \\
     & geometry & V2 & 97.57 & 19,971 \\
     & interm. algebra & V13 & 98.53 & 26,280 \\
     & number theory & V2 & 100.00 & 11,833 \\
     & prealgebra & V1 & 98.85 & 8,494 \\
     & precalculus & V8 & 97.86 & 9,173 \\
    \cmidrule(lr){1-5}
    MBPP & all & V0 & 95.85 & 9,532 \\
    \cmidrule(lr){1-5}
    \multirow{4}{*}{PlanBench} & blocks. & V0 & 100.00 & 9,581 \\
     & blocks. hard & V0 & 95.45 & 23,040 \\
     & mystery & V0 & 93.89 & 32,099 \\
     & randomized & V9 & 69.17 & 56,807 \\
    \cmidrule(lr){1-5}
    Game24 & all & V6 & 95.34 & 12,971 \\
      \bottomrule
    \end{tabular}}
    \noindent\makebox[\linewidth][c]{%
      \ifdim\wd0>\linewidth
        \resizebox{\linewidth}{!}{\box0}%
      \else
        \box0
      \fi}
  \end{minipage}%
  \hfill
  \begin{minipage}[t]{0.49\textwidth}
    \centering
    \vspace{0pt}%
    \EvalTablePanelTitle{Qwen3.6-27B}
    \setbox0=\hbox{%
    \begin{tabular}{llrrr}
      \toprule
    Benchmark & Category & Reported Version & Accuracy (\%) & Tokens/task \\
      \midrule
    \multirow{7}{*}{Math} & algebra & V3 & 99.58 & 2,071 \\
     & count./prob. & V1 & 99.30 & 11,242 \\
     & geometry & V0 & 94.79 & 9,719 \\
     & interm. algebra & V1 & 95.76 & 7,996 \\
     & number theory & V0 & 98.46 & 8,004 \\
     & prealgebra & V0 & 98.08 & 7,541 \\
     & precalculus & V0 & 96.64 & 8,343 \\
    \cmidrule(lr){1-5}
    MBPP & all & V5 & 89.12 & 1,949 \\
    \cmidrule(lr){1-5}
    \multirow{4}{*}{PlanBench} & blocks. & V7 & 96.39 & 9,151 \\
     & blocks. hard & V4 & 37.88 & 14,228 \\
     & mystery & V6 & 65.56 & 13,568 \\
     & randomized & V18 & 49.72 & 21,509 \\
    \cmidrule(lr){1-5}
    Game24 & all & V6 & 79.90 & 6,539 \\
      \bottomrule
    \end{tabular}}
    \noindent\makebox[\linewidth][c]{%
      \ifdim\wd0>\linewidth
        \resizebox{\linewidth}{!}{\box0}%
      \else
        \box0
      \fi}
  \end{minipage}%
  \endgroup
\end{table}

\subsection{Long-Horizon Evolution and Version-Level Accuracy--Token Frontiers}

Additional evolution epochs may further improve the performance reached by
EMAS. We therefore extend Game24 evolution to E15.
Table~\ref{tab:eval-game24-passes} compares the performance reached within the
two evolution horizons. For Kimi, EMAS reaches 95.34\% accuracy within E2 and
96.45\% within E15, while tokens per task decrease from 12,971 to 12,585. For
Qwen, EMAS reaches 79.90\% within E2 and 94.73\% within E15, while tokens per
task decrease from 6,539 to 3,481. Extending the Game24 evolution horizon
therefore improves both accuracy and token efficiency for both backbones. The
accuracy attained within the available evolution budget improves approximately
monotonically across the extended runs.

\begin{table}[!htbp]
  \centering
  \caption{\textbf{Long-horizon evolution further improves Game24 accuracy and token efficiency.} For each model, E2 and E15 summarize the highest accuracy EMAS reaches within the corresponding evolution budget; ties favor fewer tokens and then the earlier Version. Reported Version is formatted as Version/Epoch, indicating when the reported result was achieved. Values are averaged over three runs.}
  \label{tab:eval-game24-passes}
  \begingroup
  \EvalTableMainStyle
  \setbox0=\hbox{%
  \begin{tabular}{lllrr}
    \toprule
    Model & Endpoint & Reported Version & Accuracy (\%) & Tokens/task \\
    \midrule
    \multirow{2}{*}{Kimi} & E2 & V6/E1 & 95.34 & 12,971 \\
     & E15 & V16/E4 & 96.45 & 12,585 \\
    \cmidrule(lr){1-5}
    \multirow{2}{*}{Qwen} & E2 & V6/E2 & 79.90 & 6,539 \\
     & E15 & V19/E10 & 94.73 & 3,481 \\
    \bottomrule
  \end{tabular}}
  \noindent\makebox[\textwidth][c]{%
    \ifdim\wd0>\textwidth
      \resizebox{\textwidth}{!}{\box0}%
    \else
      \box0
    \fi}
  \endgroup
\end{table}

\begin{figure}[!htbp]
  \centering
  \includegraphics[width=\textwidth]{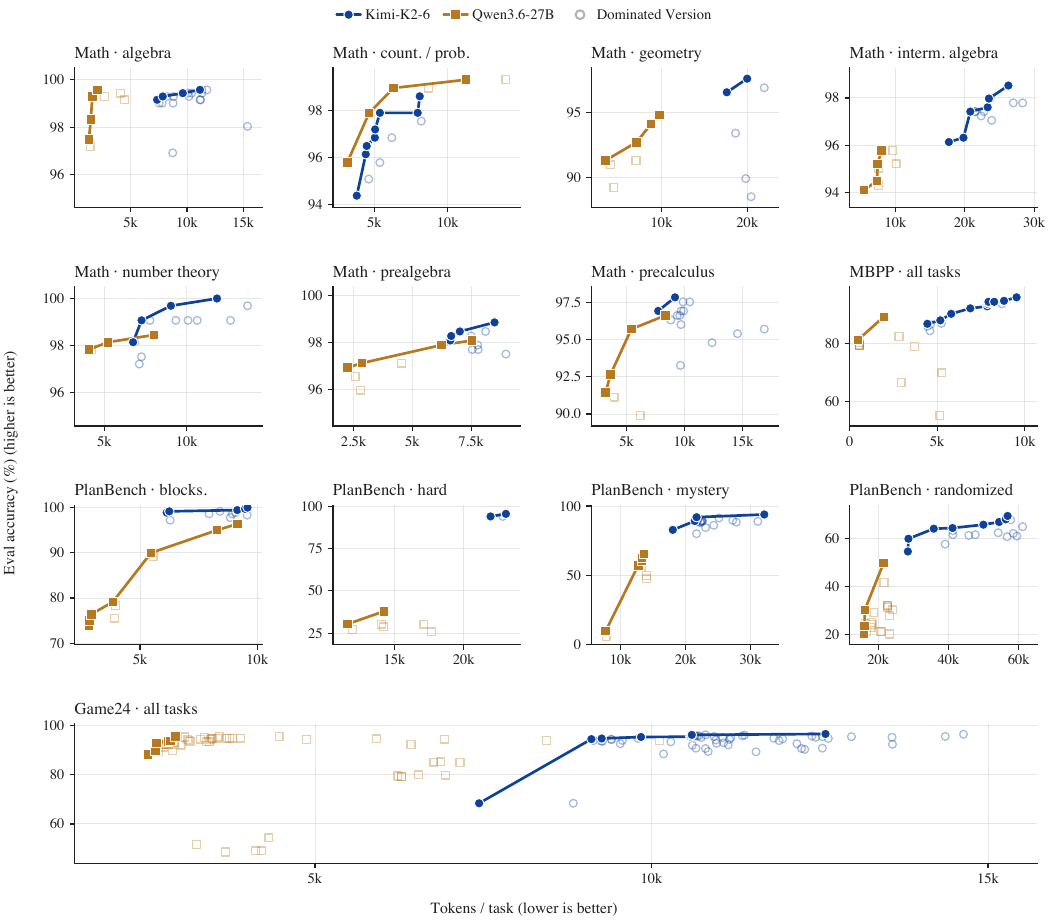}
  \input{assets/figures/eval_version_pareto_frontiers_caption.tex}
  \label{fig:appendix-version-pareto}
\end{figure}

The improvement in attained accuracy does not imply that every Version improves over its
predecessor. Figure~\ref{fig:appendix-version-pareto} shows all evaluated
Versions in accuracy--token space. Multiple categories contain more than one
Pareto-optimal Version, so the Version attaining the highest observed accuracy
is not always the Version with the lowest token cost. The non-Game24 panels include Versions through E2,
whereas the Game24 panel includes Versions through E15. The long-horizon
conclusion is therefore limited to Game24.

\FloatBarrier
\clearpage

\subsection{Revision Selectivity and Held-out Reliability}

Table~\ref{tab:eval-trigger-thresholds} lists the evidence thresholds used by
Full EMAS. These thresholds require a revision hypothesis to recur before it
can trigger an operation, providing the first stage of revision selectivity.

\begin{table}[!htbp]
  \centering
  \caption{\textbf{Operation-specific support thresholds in Full EMAS.} The threshold $\kappa_a$ is the number of distinct samples that must support the same revision hypothesis before permitted operation $a$ is triggered. The $\kappa=1$ ablation sets all thresholds to one.}
  \label{tab:eval-trigger-thresholds}
  \begingroup
  \EvalTableStandardStyle
  \setbox0=\hbox{%
  \begin{tabular}{rlr}
    \toprule
    Priority & Permitted operation & Support threshold ($\kappa_a$) \\
    \midrule
    1 & \textsc{add-node} & \multirow{3}{*}{4} \\
    2 & \textsc{split-node} &  \\
    3 & \textsc{add-edge} &  \\
    \midrule
    4 & \textsc{prompt-only} & 6 \\
    \midrule
    5 & \textsc{remove-edge} & \multirow{2}{*}{10} \\
    6 & \textsc{remove-node} &  \\
    \bottomrule
  \end{tabular}}
  \noindent\makebox[\textwidth][c]{%
    \ifdim\wd0>\textwidth
      \resizebox{\textwidth}{!}{\box0}%
    \else
      \box0
    \fi}
  \endgroup
\end{table}

\FloatBarrier

Table~\ref{tab:eval-trigger-outcomes} reports the resulting trigger outcomes
through E2. Across both backbones, 255 of 1,118 triggers are accepted. Kimi and
Qwen each produce 559 triggers, but Kimi accepts 157 (28.1\%), compared with 98
(17.5\%) for Qwen. Kimi therefore commits more revisions because more of its
triggers pass Validation, not because it produces more triggers. Because each
backbone uses its own model-conditioned Validation split, these rates are
descriptive and should not be read as a controlled comparison between
backbones.

\begin{table}[!htbp]
  \centering
  \caption{\textbf{A selective subset of trigger events becomes committed MAS revisions across two evolution epochs.} For each benchmark--category--model trajectory, the table reports the number of triggers by permitted operation and whether each trigger is ultimately accepted or rejected. Rows aggregate each trajectory from E1 through E2; the six operation columns sum to Total, Acc. and Rej. partition Total, and Acc. rate is Acc./Total. Across all trajectories, EMAS activates 1,118 triggers, of which 255 become committed revisions (22.8\%).}
  \label{tab:eval-trigger-outcomes}
  \begingroup
  \EvalTableDenseStyle
  \setbox0=\hbox{%
  \begin{tabular}{lllrrrrrrrrrr}
    \toprule
    Benchmark & Category & Model & Prompt & $-$Node & Split & $-$Edge & $+$Node & $+$Edge & Acc. & Rej. & Total & Acc. rate \\
    \midrule
    \multirow{14}{*}{Math} & \multirow{2}{*}{algebra} & Kimi-K2-6 & 42 & 1 & 5 & 0 & 0 & 0 & 18 & 30 & 48 & 37.5\% \\
     &  & Qwen3.6-27B & 21 & 32 & 0 & 0 & 0 & 0 & 7 & 46 & 53 & 13.2\% \\
    \cmidrule(lr){2-13}
     & \multirow{2}{*}{count./prob.} & Kimi-K2-6 & 21 & 1 & 0 & 0 & 0 & 0 & 11 & 11 & 22 & 50.0\% \\
     &  & Qwen3.6-27B & 15 & 7 & 0 & 0 & 0 & 0 & 5 & 17 & 22 & 22.7\% \\
    \cmidrule(lr){2-13}
     & \multirow{2}{*}{geometry} & Kimi-K2-6 & 5 & 0 & 14 & 0 & 0 & 0 & 5 & 14 & 19 & 26.3\% \\
     &  & Qwen3.6-27B & 8 & 8 & 4 & 0 & 0 & 0 & 6 & 14 & 20 & 30.0\% \\
    \cmidrule(lr){2-13}
     & \multirow{2}{*}{interm. algebra} & Kimi-K2-6 & 25 & 0 & 11 & 0 & 1 & 1 & 13 & 25 & 38 & 34.2\% \\
     &  & Qwen3.6-27B & 13 & 19 & 6 & 0 & 0 & 0 & 7 & 31 & 38 & 18.4\% \\
    \cmidrule(lr){2-13}
     & \multirow{2}{*}{number theory} & Kimi-K2-6 & 24 & 0 & 0 & 0 & 0 & 0 & 12 & 12 & 24 & 50.0\% \\
     &  & Qwen3.6-27B & 2 & 22 & 0 & 0 & 0 & 0 & 3 & 21 & 24 & 12.5\% \\
    \cmidrule(lr){2-13}
     & \multirow{2}{*}{prealgebra} & Kimi-K2-6 & 27 & 0 & 0 & 9 & 0 & 0 & 9 & 27 & 36 & 25.0\% \\
     &  & Qwen3.6-27B & 19 & 15 & 2 & 0 & 0 & 0 & 6 & 30 & 36 & 16.7\% \\
    \cmidrule(lr){2-13}
     & \multirow{2}{*}{precalculus} & Kimi-K2-6 & 16 & 4 & 2 & 0 & 0 & 0 & 13 & 9 & 22 & 59.1\% \\
     &  & Qwen3.6-27B & 7 & 3 & 12 & 0 & 0 & 0 & 5 & 17 & 22 & 22.7\% \\
    \midrule
    \multirow{2}{*}{MBPP} & \multirow{2}{*}{all} & Kimi-K2-6 & 42 & 0 & 0 & 4 & 0 & 0 & 15 & 31 & 46 & 32.6\% \\
     &  & Qwen3.6-27B & 36 & 6 & 1 & 0 & 3 & 0 & 11 & 35 & 46 & 23.9\% \\
    \midrule
    \multirow{8}{*}{PlanBench} & \multirow{2}{*}{blocks.} & Kimi-K2-6 & 70 & 8 & 0 & 9 & 0 & 0 & 12 & 75 & 87 & 13.8\% \\
     &  & Qwen3.6-27B & 4 & 46 & 4 & 0 & 6 & 0 & 9 & 51 & 60 & 15.0\% \\
    \cmidrule(lr){2-13}
     & \multirow{2}{*}{blocks. hard} & Kimi-K2-6 & 7 & 4 & 0 & 0 & 0 & 0 & 2 & 9 & 11 & 18.2\% \\
     &  & Qwen3.6-27B & 14 & 0 & 1 & 0 & 0 & 0 & 6 & 9 & 15 & 40.0\% \\
    \cmidrule(lr){2-13}
     & \multirow{2}{*}{mystery} & Kimi-K2-6 & 38 & 18 & 4 & 14 & 1 & 1 & 20 & 56 & 76 & 26.3\% \\
     &  & Qwen3.6-27B & 50 & 22 & 9 & 0 & 0 & 0 & 9 & 72 & 81 & 11.1\% \\
    \cmidrule(lr){2-13}
     & \multirow{2}{*}{randomized} & Kimi-K2-6 & 26 & 12 & 17 & 7 & 4 & 8 & 18 & 56 & 74 & 24.3\% \\
     &  & Qwen3.6-27B & 62 & 9 & 11 & 0 & 4 & 0 & 18 & 68 & 86 & 20.9\% \\
    \midrule
    \multirow{2}{*}{Game24} & \multirow{2}{*}{all} & Kimi-K2-6 & 52 & 0 & 3 & 0 & 0 & 1 & 9 & 47 & 56 & 16.1\% \\
     &  & Qwen3.6-27B & 54 & 2 & 0 & 0 & 0 & 0 & 6 & 50 & 56 & 10.7\% \\
    \midrule
    \multicolumn{3}{l}{\textbf{Kimi-K2-6 overall}} & 395 & 48 & 56 & 43 & 6 & 11 & 157 & 402 & 559 & 28.1\% \\
    \multicolumn{3}{l}{\textbf{Qwen3.6-27B overall}} & 305 & 191 & 50 & 0 & 13 & 0 & 98 & 461 & 559 & 17.5\% \\
    \cmidrule(lr){1-13}
    \multicolumn{3}{l}{\textbf{Overall}} & \textbf{700} & \textbf{239} & \textbf{106} & \textbf{43} & \textbf{19} & \textbf{11} & \textbf{255} & \textbf{863} & \textbf{1118} & \textbf{22.8\%} \\
    \bottomrule
  \end{tabular}}
  \noindent\makebox[\textwidth][c]{%
    \ifdim\wd0>\textwidth
      \resizebox{\textwidth}{!}{\box0}%
    \else
      \box0
    \fi}
  \endgroup
\end{table}

\FloatBarrier

Validation acceptance does not guarantee improvement on Test.
Table~\ref{tab:eval-transitions} examines 93 E2 revisions with an
exact-predecessor Test comparison: 64 from Kimi and 29 from Qwen. This is a
narrower scope than the 255 E1--E2 accepted revisions in
Table~\ref{tab:eval-trigger-outcomes}. Overall, 39 revisions reduce Test
accuracy, with an average loss of 3.03 percentage points. Successful
accuracy-directed revisions gain 4.46 points on average. The average gain is
also larger than the average loss within each backbone: 3.31 versus 2.54 points
for Kimi and 5.60 versus 4.29 points for Qwen. In addition, 28 of 59
cost-directed revisions reduce tokens without lowering accuracy, saving 427
tokens per task on average. Successful accuracy-directed revisions therefore
have larger mean gains than the mean losses of regressive revisions. Together
with the aggregate improvements in
Table~\ref{tab:eval-genesis-best-composite}, this pattern is consistent with
repeated evolution making overall progress despite local regressions.

\begin{table}[!htbp]
  \centering
  \caption{\textbf{Revision objectives and held-out outcomes after Validation acceptance.} For each model and benchmark, $n$ counts revisions committed during Epoch 2 and compared with their immediately preceding Versions on Test. Regressed and Avg. loss report the frequency and mean magnitude of accuracy decreases. Acc. success counts accuracy-directed revisions that improve accuracy; Cost success counts cost-directed revisions that reduce tokens without decreasing accuracy. Success cells give successful/eligible revisions (rate), and Acc. gain and Tokens saved/task are conditional means over successes.}
  \label{tab:eval-transitions}
  \begingroup
  \EvalTableDenseStyle
  \setbox0=\hbox{%
  \begin{tabular}{llrrrrrrr}
    \toprule
    Model & Benchmark & $n$ & Regressed $n$ (\%) & Avg. loss (pp) & Acc. success & Acc. gain (pp) & Cost success & Tokens saved/task \\
    \midrule
    \textbf{Overall} & \textbf{All} & \textbf{93} & \textbf{39 (41.9\%)} & \textbf{3.03} & \textbf{20/34 (58.8\%)} & \textbf{4.46} & \textbf{28/59 (47.5\%)} & \textbf{427} \\
    \midrule
    \multirow{5}{*}{\textbf{Kimi-K2-6}} & \textbf{All} & \textbf{64} & \textbf{28 (43.8\%)} & \textbf{2.54} & \textbf{10/17 (58.8\%)} & \textbf{3.31} & \textbf{22/47 (46.8\%)} & \textbf{465} \\
     & Math & 30 & 10 (33.3\%) & 1.08 & 4/7 (57.1\%) & 1.80 & 13/23 (56.5\%) & 634 \\
     & MBPP & 8 & 5 (62.5\%) & 2.80 & 1/2 (50.0\%) & 2.59 & 2/6 (33.3\%) & 318 \\
     & PlanBench & 23 & 11 (47.8\%) & 3.67 & 5/7 (71.4\%) & 4.67 & 7/16 (43.8\%) & 195 \\
     & Game24 & 3 & 2 (66.7\%) & 2.94 & 0/1 (0.0\%) & -- & 0/2 (0.0\%) & -- \\
    \midrule
    \multirow{5}{*}{\textbf{Qwen3.6-27B}} & \textbf{All} & \textbf{29} & \textbf{11 (37.9\%)} & \textbf{4.29} & \textbf{10/17 (58.8\%)} & \textbf{5.60} & \textbf{6/12 (50.0\%)} & \textbf{286} \\
     & Math & 7 & 3 (42.9\%) & 2.26 & 1/3 (33.3\%) & 4.59 & 2/4 (50.0\%) & 274 \\
     & MBPP & 4 & 1 (25.0\%) & 0.52 & -- & -- & 3/4 (75.0\%) & 32 \\
     & PlanBench & 17 & 7 (41.2\%) & 5.69 & 8/13 (61.5\%) & 6.34 & 1/4 (25.0\%) & 1,072 \\
     & Game24 & 1 & 0 (0.0\%) & -- & 1/1 (100.0\%) & 0.74 & -- & -- \\
    \bottomrule
  \end{tabular}}
  \noindent\makebox[\textwidth][c]{%
    \ifdim\wd0>\textwidth
      \resizebox{\textwidth}{!}{\box0}%
    \else
      \box0
    \fi}
  \endgroup
\end{table}

\FloatBarrier

Table~\ref{tab:eval-headroom-alignment} groups the same 93 revisions by their
starting accuracy. Panel A uses V0 accuracy as global headroom, while Panel B
uses the exact predecessor as local headroom. Among successful
accuracy-directed revisions, those starting below 50\% predecessor accuracy
gain 6.88 points on average, compared with 1.35 points for those with saturated
predecessors. However, regressions occur in every bin, and the overall success
rate does not vary monotonically with headroom. Greater accuracy headroom is
therefore associated with larger gains when accuracy revisions succeed, but
does not by itself determine whether a Validation-accepted revision will
transfer successfully.

\begin{table}[!htbp]
  \centering
  \caption{\textbf{Accuracy headroom and held-out outcomes of committed revisions.} The 93 Epoch-2 revisions are grouped by V0 accuracy in Panel A and by preceding-Version accuracy in Panel B; lower accuracy indicates greater headroom. Regressed counts revisions that lower Test accuracy, and Successful counts revisions that improve accuracy or reduce tokens without lowering accuracy. Losses, gains, and token savings are averaged over the corresponding revisions.}
  \label{tab:eval-headroom-alignment}
  \begingroup
  \EvalTableDenseStyle
  \EvalTablePanelTitle{Panel A: V0 accuracy (global headroom)}
  \setbox0=\hbox{%
  \begin{tabular}{lrrrrrr}
    \toprule
    V0 accuracy bin & $n$ & Regressed $n$ (\%) & Avg. loss (pp) & Successful $n$ (\%) & Acc. gain (pp) & Tokens saved/task \\
    \midrule
    Low ($<50$) & 15 & 6 (40.0\%) & 5.81 & 8 (53.3\%) & 6.88 & 1,072 \\
    Medium ($50$--$80$) & 19 & 9 (47.4\%) & 4.32 & 9 (47.4\%) & 3.32 & 32 \\
    High ($80$--$95$) & 2 & 1 (50.0\%) & 1.04 & 1 (50.0\%) & -- & 372 \\
    Saturated ($\geq95$) & 57 & 23 (40.4\%) & 1.89 & 30 (52.6\%) & 3.01 & 453 \\
    \bottomrule
  \end{tabular}}
  \noindent\makebox[\textwidth][c]{%
    \ifdim\wd0>\textwidth
      \resizebox{\textwidth}{!}{\box0}%
    \else
      \box0
    \fi}
  \EvalTablePanelTitle{Panel B: exact-predecessor accuracy (local headroom)}
  \setbox0=\hbox{%
  \begin{tabular}{lrrrrrr}
    \toprule
    Predecessor bin & $n$ & Regressed $n$ (\%) & Avg. loss (pp) & Successful $n$ (\%) & Acc. gain (pp) & Tokens saved/task \\
    \midrule
    Low ($<50$) & 12 & 4 (33.3\%) & 5.80 & 8 (66.7\%) & 6.88 & 1,072 \\
    Medium ($50$--$80$) & 17 & 8 (47.1\%) & 4.96 & 8 (47.1\%) & 3.48 & 32 \\
    High ($80$--$95$) & 19 & 10 (52.6\%) & 2.11 & 9 (47.4\%) & 3.90 & 705 \\
    Saturated ($\geq95$) & 45 & 17 (37.8\%) & 2.02 & 23 (51.1\%) & 1.35 & 398 \\
    \bottomrule
  \end{tabular}}
  \noindent\makebox[\textwidth][c]{%
    \ifdim\wd0>\textwidth
      \resizebox{\textwidth}{!}{\box0}%
    \else
      \box0
    \fi}
  \endgroup
\end{table}

\FloatBarrier

\subsection{Evolution Budget Through Epoch 2}

Table~\ref{tab:eval-evolution-cost} reports the directly recorded evolution
execution cost through E2. Across the two backbones and four benchmarks, the
measured cost totals 2.929B tokens. Kimi accounts for 2.063B tokens and Qwen
for 866M tokens, with Kimi higher on every benchmark.

This difference is consistent with Table~\ref{tab:eval-main}. Kimi uses more
tokens per task than Qwen for both the Initial MAS (14.81k versus 7.44k
overall) and EMAS (17.07k versus 8.01k overall), with the same ordering on all
four benchmarks. The higher Kimi evolution total is therefore consistent with
its higher observed per-task token scale, although
Table~\ref{tab:eval-evolution-cost} does not isolate this factor from
differences in execution and revision behavior.

Across benchmarks, PlanBench has the largest measured cost, followed by Math,
Game24, and MBPP. Evolution cost is therefore concentrated rather than uniform
across settings. These totals cover only the directly recorded E1--E2
execution tokens; they are not a complete end-to-end cost or a controlled
comparison of model efficiency.

\begin{table}[!htbp]
  \centering
  \caption{\textbf{Total measured evolution cost across models and benchmarks.} Costs are reported through Epoch 2.}
  \label{tab:eval-evolution-cost}
  \begingroup
  \EvalTableStandardStyle
  \setbox0=\hbox{%
  \begin{tabular}{llr}
    \toprule
    Model & Benchmark & Evolution cost (tokens) \\
    \midrule
    \textbf{Overall} & \textbf{All} & \textbf{2.929B} \\
    \cmidrule(lr){1-3}
    \multirow{4}{*}{Kimi-K2-6} & Math & 762.37M \\
     & MBPP & 58.81M \\
     & PlanBench & 918.23M \\
     & Game24 & 323.41M \\
    \cmidrule(lr){1-3}
    \multirow{4}{*}{Qwen3.6-27B} & Math & 178.98M \\
     & MBPP & 11.56M \\
     & PlanBench & 556.78M \\
     & Game24 & 118.76M \\
    \bottomrule
  \end{tabular}}
  \noindent\makebox[\textwidth][c]{%
    \ifdim\wd0>\textwidth
      \resizebox{\textwidth}{!}{\box0}%
    \else
      \box0
    \fi}
  \endgroup
\end{table}

\FloatBarrier

\subsection{Ablation Stability: Aggregate Summary and Full Trajectories}

\begin{figure}[!htbp]
  \centering
  \includegraphics[width=\textwidth]{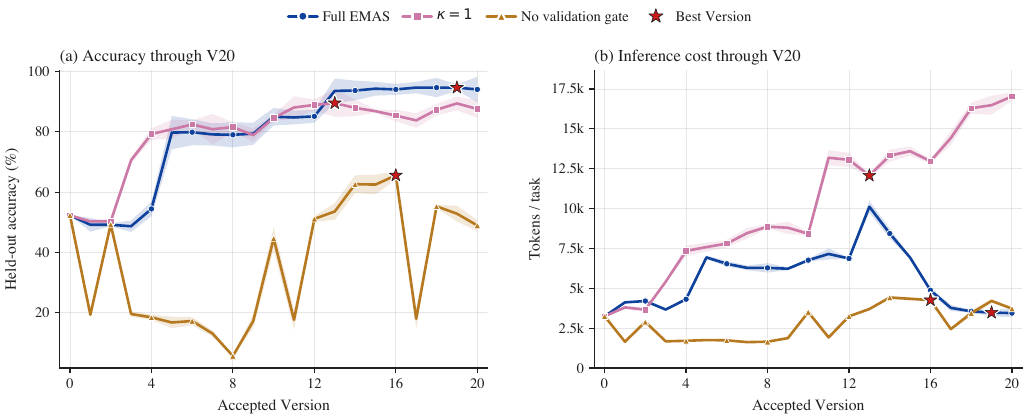}
  \input{assets/figures/eval_ablation_complete_caption.tex}
  \label{fig:eval-ablation-complete}
\end{figure}

Table~\ref{tab:eval-ablation-stability-detail} and
Figure~\ref{fig:eval-ablation-complete} expand the ablation summary in
Table~\ref{tab:eval-core-ablations}. They show where the best checkpoint occurs
and how each setting evolves from V0 to V20, rather than only the aggregate
result.

Full EMAS reaches its best checkpoint at V19, with 94.73\% accuracy and 3,481
tokens per task. Seven of its 20 transitions reduce accuracy, but these losses
sum to only 5.59 percentage points. Stability therefore does not mean that
every accepted revision improves. Instead, the regressions remain small enough
that later Versions often recover and reach higher accuracy.

With $\kappa=1$, the best checkpoint appears earlier at V13, with 89.58\%
accuracy and 12,075 tokens per task. Although it has only one more regressive
transition than Full EMAS (eight versus seven), its cumulative loss increases
to 13.92 points and its token cost grows more. In this comparison, evidence
accumulation is therefore associated with smaller cumulative damage and lower
token growth, rather than the removal of every downward step. Without the
Validation gate, 11 of 20 transitions regress, total regression rises to
158.53 points, and even the best checkpoint reaches only 65.56\% accuracy.
Removing the gate produces the largest instability among the three settings.

\begin{table}[!htbp]
  \centering
  \caption{\textbf{Ablation settings and evolution stability through V20.} Results use Qwen3.6-27B on Game24. Best Version is the highest-accuracy checkpoint, with fewer tokens and then the earlier Version breaking ties. The Regressive steps column counts accuracy decreases from the preceding Version. Total regression sums these decreases, and Avg. regression averages them over all 20 transitions. Values are averaged over three runs.}
  \label{tab:eval-ablation-stability-detail}
  \begingroup
  \EvalTableStandardStyle
  \setbox0=\hbox{%
  \begin{tabular}{lrrrrrr}
    \toprule
    Ablation & Best Version & Best acc. (\%) & Tokens/task & Regressive steps & Total regression (pp) & Avg. regression (pp) \\
    \midrule
    Full EMAS & V19 & 94.73 & 3,481 & 7/20 & 5.59 & 0.28 \\
    $\kappa=1$ & V13 & 89.58 & 12,075 & 8/20 & 13.92 & 0.70 \\
    No gate & V16 & 65.56 & 4,262 & 11/20 & 158.53 & 7.93 \\
    \bottomrule
  \end{tabular}}
  \noindent\makebox[\textwidth][c]{%
    \ifdim\wd0>\textwidth
      \resizebox{\textwidth}{!}{\box0}%
    \else
      \box0
    \fi}
  \endgroup
\end{table}

\FloatBarrier

Figure~\ref{fig:eval-ablation-complete} makes the trajectory differences
visible. Full EMAS improves and remains at high accuracy across the later
Versions, while its token cost falls again after a temporary increase. The
$\kappa=1$ setting improves early but becomes less stable as its token cost
grows. Without the gate, accuracy repeatedly collapses and recovers. The
advantage reported in Table~\ref{tab:eval-core-ablations} therefore reflects
the broader trajectory rather than a single favorable checkpoint.

\FloatBarrier

Together, these results show that the two controls improve different aspects
of stability in this Qwen3.6-27B Game24 study. Evidence accumulation is
associated with smaller cumulative regression and lower token growth, while
removing the Validation gate produces much larger held-out regressions. Neither
control makes evolution monotonic, and the conclusion is limited to this
ablation setting.

\end{document}